\documentclass[sigconf, screen]{acmart}
\AtBeginDocument{%
  \providecommand\BibTeX{{%
    \normalfont B\kern-0.5em{\scshape i\kern-0.25em b}\kern-0.8em\TeX}}}

\copyrightyear{2023}
\acmYear{2023}
\setcopyright{acmlicensed}\acmConference[ICAIF '23]{4th ACM International Conference on AI in Finance}{November 27--29, 2023}{Brooklyn, NY, USA}
\acmBooktitle{4th ACM International Conference on AI in Finance (ICAIF '23), November 27--29, 2023, Brooklyn, NY, USA}
\acmPrice{15.00}
\acmDOI{10.1145/3604237.3626891}
\acmISBN{979-8-4007-0240-2/23/11}

\newcommand{\stdsign}{$\pm$\xspace}

\usepackage{enumitem}

\usepackage{xspace}
\usepackage{textcomp} 
\usepackage{float}
\usepackage{listings}
\usepackage{caption}
\usepackage{upgreek}
\usepackage{booktabs}
\usepackage{multirow}
\usepackage{array}
\usepackage{multicol}
\usepackage{fancybox}

\newcommand{\microf}{$\mathrm{\upmu}$-$\mathrm{F_1}$\xspace}

\newcommand{\macrof}{$\mathrm{m}$-$\mathrm{F_1}$\xspace}

\begin{document}

\title{Making LLMs Worth Every Penny: \\ Resource-Limited Text Classification in Banking}


\author{Lefteris Loukas}
\email{lefteris.loukas@helvia.ai}
\affiliation{%
    \institution{Helvia.ai}
    \institution{Dept. of Informatics, Athens Univ. of Economics and Business, Greece}
    \streetaddress{}
    \city{}
    \country{}
    \postcode{}
}

\author{Ilias Stogiannidis}
\email{ilias.stogiannidis@helvia.ai}
\affiliation{%
\institution{Helvia.ai}
\institution{Dept. of Informatics, Athens Univ. of Economics and Business, Greece}
\streetaddress{}
\city{}
\country{}
\postcode{}}

\author{Odysseas Diamantopoulos}
\email{odysseas.diamantopoulos@helvia.ai}
\affiliation{%
\institution{Helvia.ai}
\institution{Dept. of Informatics, Athens Univ. of Economics and Business, Greece}
\streetaddress{}
\city{}
\country{}
\postcode{}}

\author{Prodromos Malakasiotis}
\affiliation{%
\institution{Dept. of Informatics, Athens Univ. of Economics and Business, Greece}
\streetaddress{}
\city{}
\country{}
\postcode{}}

\author{Stavros Vassos}
\email{stavros@helvia.ai}
\affiliation{%
\institution{Helvia.ai}
\streetaddress{}
\city{}
\country{}
\postcode{}}

\renewcommand{\shortauthors}{Loukas et al.}

\begin{abstract}
Standard Full-Data classifiers in NLP demand thousands of labeled examples, which is impractical in data-limited domains. 
Few-shot methods offer an alternative, utilizing contrastive learning techniques that can be effective with as little as 20 examples per class. Similarly, Large Language Models (LLMs) like GPT-4 can perform effectively with just 1-5 examples per class. However, the performance-cost trade-offs of these methods remain underexplored, a critical concern for budget-limited organizations. Our work addresses this gap by studying the aforementioned approaches over the Banking77 financial intent detection dataset, including the evaluation of cutting-edge LLMs by OpenAI, Cohere, and Anthropic in a comprehensive set of few-shot scenarios. We complete the picture with two additional methods: 
first, a cost-effective querying method for LLMs based on retrieval-augmented generation (RAG), able to reduce operational costs multiple times compared to classic few-shot approaches, and second, a data augmentation method using GPT-4, able to improve performance in data-limited scenarios.
Finally, to inspire future research, 
we provide a human expert's curated subset of Banking77, along with extensive error analysis.
\end{abstract}






\maketitle

\begin{table}[t]
    \centering
\small
\begin{tabular}{lc}
        \toprule
        \textbf{Intent}  & \textbf{Label} \\
        \midrule

My card was declined. & Declined Card Payment \\
It declined my transfer. & Declined Transfer \\
How do you calculate your exchange rates? & Exchange Rate \\
My card was eaten by the cash machine. & Card Swallowed \\
I lost my card in the ATM. & Card Swallowed \\
I got married, I need to change my name. & Edit Personal Details \\

... & ... \\
My card is needed soon. & Card Delivery Estimate\\
What is the youngest age for an account?	& Age Limit \\

\bottomrule
    \end{tabular}
    \caption{Example banking intents and their labels from the Banking77 dataset. In total, there are 77 different labels in the dataset with highly overlapping semantic similarities.}
    \label{tab:example-of-banking77}
    \vspace{-8mm}
\end{table}

\section{Introduction}

The field of Natural Language Processing (NLP) has seen impressive advancements in the past few years, with particular emphasis on text classification. 
Traditional full-data classifiers require thousands of labeled samples, making them infeasible for data-limited domains such as finance \cite{banking77-casanueva-etal-2020-efficient}. 
Modern Few-Shot techniques, which include contrastive learning \cite{setfit-Tunstall2022EfficientFL}, aim to alleviate this issue by requiring only 10 to 20 examples per class. 
Also, recent advancements focus on prompting large Language Models (LLMs) like GPT-3 with as few as 1-5 examples per class, typically via a managed API.
However, the tradeoffs concerning the performance and Operating Costs (OpEx) of available methods remain under-explored. 

In this paper, we bridge this gap by evaluating the aforementioned approaches in a comprehensive set of few-shot scenarios over a financial intent classification dataset, Banking77 \cite{banking77-casanueva-etal-2020-efficient}, including the evaluation of cutting-edge LLMs by OpenAI, Cohere, and Anthropic. 
Banking77 is a real-life dataset containing customer service intents and their classification labels. Contrary to other intent detection datasets, Banking77 contains a large number (77) of labels with semantic overlaps (Table ~\ref{tab:example-of-banking77}). 
These characteristics make it suitable for investigating methodological perspectives while, at the same time, solving a business use case.

First, we fine-tune MPNet \cite{mpnet-2020}, a pre-trained Masked Language Model (MLM) by providing it with the complete dataset (Full-Data Setting with $\sim$10k training examples). 
Second, we fine-tune MPNet again, but this time, we use SetFit \cite{setfit-Tunstall2022EfficientFL}, a contrastive learning technique that can achieve comparable results when shown only up to 20 examples per class (Few-Shot), instead of the complete dataset. 
Third, we leverage in-context learning with a wide range of popular conversational LLMs, including GPT-3.5, GPT-4, and Cohere's Command-nightly and Anthropic's Claude 2. We provide the LLMs with only 1 and 3 examples per class (Few-Shot), which we assume is practical from a business perspective in data-limited domains like finance \cite{loukas-etal-2021-edgar}. In this setting, we also demonstrate that curated samples picked by a hired human expert outperform randomly selected ones with up to 10 points difference.
Overall, we show that in-context learning with LLMs can outperform fine-tuned masked language models (MLMs) in financial few-shot text classification tasks, even when presented with fewer examples. 
Apart from studying performance, we also present a cost analysis on employing LLMs. Motivated by this, we extend our study with a cost-effective LLM inference method based on Retrieval-Augmented Generation (RAG) that reduces costs significantly, enabling small organizations to leverage LLMs without incurring substantial expenses. We demonstrate that with this approach, we retrieve only a tiny but crucial fraction (2.2\%) of examples compared to the classic few-shot settings, and we are able to surpass the performance of the most competitive method (GPT-4) by 1.5 points while costing 700\$ less in the context of the test set ($\sim$3k examples). Furthermore, we simulate a low-resource data scenario, where we augment the training dataset by leveraging GPT-4, demonstrating improved performance. We also analyze the threshold where data augmentation loses effectiveness, aiding AI practitioners in their decision-making. 

To the best of our knowledge, this is the first study that presents together a comprehensive evaluation of such methods 
in a resource-limited industrial context, where data and budget availability are key factors.

\section{Related Work}

\subsection{Studies on Banking77}
Casanueva et al. \shortcite{banking77-casanueva-etal-2020-efficient} introduced Banking77 and achieved baseline accuracy scores of 93.6\% and 85.1\% for Full-Data and 10-shot settings by fine-tuning BERT \cite{devlin-etal-2019-bert} and using Universal Sentence Encoders \cite{use-1-cer-etal-2018-universal}. 
Ying and Thomas \shortcite{ying-thomas-2022-label-banking77-1} later improved these results by addressing label errors in the dataset, utilizing confident learning and cosine similarity approaches for mislabeled utterance detection ~\cite{northcutt2021confidentlearning}. Their trimmed dataset yielded a significant performance boost, with a 92.4\% accuracy and 92.0\% F1 score. Li et al. \shortcite{li-etal-2022-learning-better} demonstrated that pre-training intent representations can improve performance in financial intent classification. They achieved an 82.76\% accuracy and 87.3\% Macro-F1 score on the Banking77 Full-Data benchmark using prefix-tuning and fine-tuning of the last LLM layer. 
Lastly, Mehri and Eric \shortcite{mehri-eric-2021-example} proposed two dialogue system text classification approaches: "observers" and example-driven training. Observers offered an alternative to the [CLS] token for semantic representation and their example-driven training method leveraged sentence similarity for classification. They achieved an accuracy of 85.9\% in the 10-shot setting and 93.8\% in the Full-Data setting.

\subsection{Few-Shot Text Classification}

Learning from only a few training instances is crucial, especially in real-world use cases where there is no prior dataset and typically there are limited or no resources to create one. In such cases with very limited data, fine-tuning often performs poorly \cite{dodge2020finetuning} and actually becoming more challenging as language models grow in size. An alternative approach is in-context learning \cite{brown2020language-gpt3}, which involves prompting a generative large language model (LLM) with a context and asking it to complete NLP tasks without fine-tuning. The context usually includes a brief task description, some examples (the context), and the instance to be classified. The idea behind in-context learning is that the language model has already learned several tasks during pre-training, and the prompt attempts to identify the appropriate one \cite{reynolds2021}. However, selecting the right prompt is not easy as language models cannot understand the meaning of the prompt \cite{webson-pavlick-2022-prompt}. To address this issue, LLMs have been fine-tuned to follow human instructions \cite{instruct-gpt,openai2023-technical-report-gpt4}. Despite this improvement, in-context learning is still correlated by term frequencies encountered during pre-training \cite{razeghi-etal-2022-impact}. At the same time, instruct-tuned LLMs (like GPT-3.5 and GPT-4 by OpenAI), carry the biases of the human annotators who provided the training instructions. To overcome these challenges, prompt-tuning has emerged as a promising research direction \cite{prompt-tuning-1-lester-etal-2021-power,prompt-tuning-2-jiaetal-2022}.

\section{Task and Dataset}

\begin{table}[t]
\centering

    \begin{tabular}{lcc}
    \toprule
    \textbf{Banking77 Statistics} & \textbf{Train} & \textbf{Test} \\
    \midrule
        Number of examples & 10,003 & 3,080 \\
        \midrule
        Average length (in characters)  & 59.5 \stdsign 40.9 & 54.2 \stdsign 34.7 \\
        \midrule
        Average length (in words) & 11.9 \stdsign 7.9 & 10.9 \stdsign 6.7\\
        \midrule
        Number of intents (classes) & 77 & 77 \\
    \bottomrule
    \end{tabular}
    \caption{Banking77 dataset statistics. The average lengths are shown along with their corresponding standard deviations.} 
    \label{tab:banking77-stats}
    \vspace{-6mm}
    \end{table}

Intent detection is a particular case of text classification, and it is a vital component of task-oriented conversational systems in various domains, including finance. It reflects the complexity of actual commercial systems, which can be attributed to the partially overlapping intent categories, the need for fine-grained decisions, and the  lack of comprehensive datasets in finance \cite{banking77-casanueva-etal-2020-efficient,loukas-etal-2021-edgar,loukas-etal-2022-finer,zavitsanos-etal-2021}. 

However, publicly available intent detection datasets are limited, and the existing ones fail to represent the complexity of real-world industrial systems \cite{simple-intent-datasets-1-braun-etal-2017-evaluating,simple-intents-coucke2018snips}. In response to the need for industry-ready datasets \cite{better-intents-3-larson-etal-2019-evaluation,better-intent-2b-Liu2021}, PolyAI released Banking77 \cite{banking77-casanueva-etal-2020-efficient}, which focuses on a single domain and consists of 77 fine-grained intents related to banking. By concentrating on a specific domain and offering a diverse set of intents, the dataset emulates a more realistic and challenging intent detection task than most generic benchmarks. Also, it is worth noting that some intent categories partially overlap, requiring fine-grained decisions that cannot rely solely on the semantics of individual words, further highlighting the task's difficulty.

\begin{figure}[h]
  \centering
  \includegraphics[width=0.7\columnwidth]{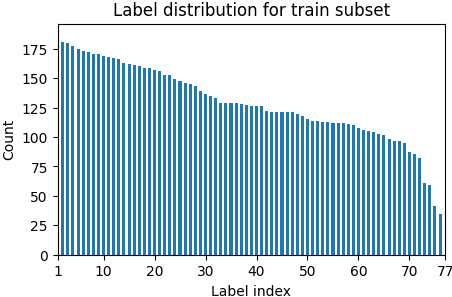}
  \caption{Class distribution of the 77 intents in the training set. Intent indices are shown instead of tag names for brevity.}
  \label{fig:train_subset_label_distribution}
  \vspace{-1mm}
\end{figure}

The dataset comprises 13,083 annotated customer service queries labeled with 77 intents and is split into two subsets: train (10,003 examples) and test (3,080 samples) (Table ~\ref{tab:banking77-stats}). 
The label distribution is heavily imbalanced in the training subset (Figure \ref{fig:train_subset_label_distribution}), demonstrating the challenge of developing classifiers in the Full-Data setting.\footnote{The test subset comprises 40 instances for every label.}\footnote{This paper is an extended version of our previous work \citep{loukas-etal-2023-breaking-the-bank-finnlp}.}

\section{Methodology}

\subsection{Fine-tuning MLMs}

\textbf{MPNet} \cite{mpnet-2020} is a transformer-based model \cite{Vaswani2017AttentionIA-transformer,devlin-etal-2019-bert} designed with a unique pre-training objective, employing permuted language modeling to capture token dependencies and utilizing auxiliary position information as input. Pre-trained on a substantial 160GB text corpora, MPNet demonstrates superior performance compared to BERT \cite{devlin-etal-2019-bert}, XLNet \cite{xlnet-2019}, and RoBERTa \cite{Liu2019RoBERTaAR} across various downstream tasks. We utilize \texttt{all-mpnet-base-v2}\footnote{\url{https://huggingface.co/sentence-transformers/all-mpnet-base-v2}}, a top-performing variation of MPNet in the sentence transformers leaderboard, making it a prominent choice for our task.\footnote{\url{https://www.sbert.net/docs/pretrained_models.html}} 

\subsection{Few-Shot Contrastive Learning with SetFit}

\textbf{SetFit} \cite{setfit-Tunstall2022EfficientFL} is a recent methodology developed by HuggingFace which fine-tunes a Sentence Transformer model on a minimal number of labeled text pairs for each class.\footnote{\url{https://github.com/huggingface/setfit}}
SetFit utilizes contrastive learning \cite{hinton-contrastivelearning-2020} in a Siamese manner, where positive and negative pairs are constructed by in- and out-class selection. As a result, transformers using SetFit produce highly descriptive text representations, where a classification head is later trained on. Despite using limited training data (such as eight training examples per class), \citeauthor{setfit-Tunstall2022EfficientFL} showed that SetFit's performance is comparable to models trained on complete datasets with standard fine-tuning \cite{setfit-Tunstall2022EfficientFL}.

\subsection{In-Context Learning}
For in-context learning, we use closed-source LLMs like \textbf{GPT-3.5} \cite{instruct-gpt} and \textbf{GPT-4} \cite{openai2023-technical-report-gpt4}, which are based on the Generative Pre-trained Transformer (GPT) \cite{radford2018improving-gpt1,radford2019language-gpt2}. These pre-trained models are further trained to follow instructions with Reinforcement Learning from Human Preferences (RLHF) \cite{NIPS2017_d5e2c0ad-DRLHP,ouyang2022training}.
GPT-3.5 is a 175B-parameter model and, at the time of the writing, has 2 variants able to consume a context of 4,096 and 16,384 tokens. GPT-4 is a multi-modal model with 2 variants available, able to consume 8,192 and 32,768 tokens.
We also used \textbf{Cohere's Command-nightly}  with a 4,096 tokens context window and \textbf{Anthropic Claude 2}, which provides a 100,000 tokens context window.\footnote{See \url{https://cohere.com/} and \url{https://www.anthropic.com/}. We use the Command-nightly version available during the week of 2-9 October 2023. Anthropic does not specify their checkpoints for the Claude 2 model.}

\subsection{Human Expert Annotation for Robustness}

\citeauthor{banking77-casanueva-etal-2020-efficient} \cite{banking77-casanueva-etal-2020-efficient} discovered class overlaps in creating Banking77. To address this issue, previous studies like \citeauthor{ying-thomas-2022-label-banking77-1} \cite{ying-thomas-2022-label-banking77-1} used additional annotation to curate subsets and enhance performance on real-life noisy datasets like Banking77. However, they did not share their curated subset for reproducibility. 

To tackle these challenges, we curated a subset of Banking77 by hiring a subject matter expert. We provided the human expert with 10, randomly-picked examples per class, and they selected the top 3 based on their alignment with the corresponding intents. This meticulous approach reduced class overlaps and ensured the high relevance of each example to its intended label. As we show later, this is fundamental in the few-shot scenario since these expert-selected training instances outperform randomly selected instances per class. To support robust financial research, we provide this curated subset as a free resource for the financial AI community.\footnote{The curated subset is hosted at \url{https://anonymous.4open.science/r/data-prompt-4805}}

\section{Experiments \& Results}
\subsection{Experimental Setup}


For the \textbf{fine-tuning} methods, we use TensorFlow and HuggingFace. For all \textbf{in-context learning} methods, we instruct the model to return only the financial intent label of the test example that is presented in each query.
The prompt we use can be broken down into 3 parts and is the same for all LLMs, i.e., GPT-3.5 and GPT-4 (OpenAI),\footnote{We use the available \texttt{gpt-3.5-turbo} variants of 4K and 16K contexts for the 1-shot and 3-shot settings accordingly, and the \texttt{gpt-4} 8K context model, which is less expensive than the 32K one. In all cases, we use the \texttt{0613} checkpoints.} Command (Cohere), and Claude (Anthropic).
The first part contains the description of the task and the available classes. The second provides a few examples, and the third presents the text to be classified. The prompt template is shown in Figure \ref{fig:prompt_template}.





\begin{figure}[ht]
\scriptsize
\setlength{\fboxrule}{0.6pt}
\setlength{\fboxsep}{8pt}
\begin{Sbox}
\begin{minipage}{0.36\textwidth}
\begin{verbatim}
You are an expert assistant in the field of customer service. 
Your task is to help workers in the customer service 
department of a company. Your task is to classify the 
customer's question in  order to help the customer service 
worker to answer the question.

In order to help the worker, you MUST respond 
with the number and the name of one of the
following classes you know. If you cannot answer the question, 
respond: "-1 Unknown".  

In case you reply with something else, you will be penalized.

The classes are: 
0 activate_my_card
1 age_limit
2 apple_pay_or_google_pay
3 atm_support
..       ..
75 wrong_amount_of_cash_received
76 wrong_exchange_rate_for_cash_withdrawal

Here are some examples of questions and their classes:
How do I top-up while traveling? automatic_top_up
How do I set up auto top-up? automatic_top_up
...               ... 
It declined my transfer. declined_transfer
\end{verbatim}
\end{minipage}
\end{Sbox}
\fbox{\TheSbox}
\caption{The prompt template used to query the Large Language Models (LLMs).}
\label{fig:prompt_template}
\end{figure}
\newpage

\subsection{Hyperparameter tuning in MLMs}
We tuned the MLM (MPNet-v2) using Optuna's \cite{optuna2019} implementation of the Tree-structured Parzen Estimator (TPE) algorithm \cite{tpse2011bergstra}. We specified 10 trials, and we defined a search space of (1e-5, 5e-5) for the body's learning rate and (1e-2, 5e-5) for the head's learning rate with logarithmic intervals. During tuning, we maximized the validation Micro-F1. We deployed the MLM pipelines in an NVIDIA A100 SXM GPU with 40GB memory.

\subsection{Prompt Engineering in LLMs}\label{subsec:prompt_engineering_subsec}
We experiment with two different prompt settings using GPT-4 in a 3-shot setting on a held-out validation subset, which was created by using 5\% of the training subset.
In the first setting, we include the few-shot examples as part of the chat history, i.e., the query is presented as a user message and the class is presented as an assistant message that follows. In the second setting, we include the few-shot examples in the so-called \texttt{system}, which is intended to give an initial context to the assistant.
The second setting yielded the best results (Table~\ref{tab:prompt_engineering}), and we use it for the rest of our experiments.

\begin{table}[h]
\centering

\begin{tabular}{@{}l|cc@{}}

\toprule
Few-shot examples given as & \microf & \macrof \\ \midrule
Previous chat history & 75.5 &  74.4\\
System context & \textbf{77.7} & \textbf{77.0}  \\
\bottomrule
\end{tabular} 

\caption{Validation Micro-F1 and Macro-F1 scores for our two prompt settings with GPT-4 in the 3-Shot scenario.} 
\label{tab:prompt_engineering} 
\end{table}


\vspace{-8mm}\subsection{Results}\label{sec:results}

To comprehensively understand how accurately the models can perform, we report micro-F1 (\microf) and macro-F1 (\macrof). 
Table \ref{tab:classification_results} shows MPNet-v2 achieving competitive results across all few-shot settings using SetFit. 
When trained on only 3 samples, it achieves scores of 76.7 \microf and 75.9 \macrof. 
As we increase the number of samples, the performance improves, reaching a 91.2 micro-F1 and 91.3 macro-F1 score with 20 samples. 
This is only 3 percentage points (pp) lower than fine-tuning the model with all the data, demonstrating the effectiveness of SetFit, especially in domains where acquiring data points is difficult.
Lastly, our MPNet-v2 solution outperforms the previous results from \cite{mehri-eric-2021-example}, both in the 10-shot and Full-Data settings (by 2.2 pp and 0.3pp).

\begin{table}[t]
\centering
\small
    \begin{tabular}{lccc}
    \toprule
        Methods & Setting & \microf & \macrof \\
            \midrule
                Mehri and Eric \shortcite{mehri-eric-2021-example} & Full-Data & 93.8 & NA \\
                Mehri and Eric \shortcite{mehri-eric-2021-example} & 10-shot & 85.9 & NA \\
                Ying and Thomas \shortcite{ying-thomas-2022-label-banking77-1} & Full-Data & NA & 92.0 \\
            \midrule
                MPNet-v2 & Full-Data & \textbf{94.1} & \textbf{94.1} \\
            \midrule
                MPNet-v2 (SetFit) & 1-shot & 57.4 & 55.9 \\
                
                GPT-3.5 (representative samples) & 1-shot & 75.2 & 74.3 \\
                GPT-3.5 (random samples) & 1-shot & 74.0 & 72.3 \\
                GPT-4 (representative samples) & 1-shot & \textbf{80.4} & \textbf{78.1} \\
                GPT-4 (random samples) & 1-shot & 77.6 & 76.7 \\
                Command-nightly (representative samples) & 1-shot & 58.4 & 57.8 \\
                Anthropic Claude 1 (representative samples) & 1-shot & 73.8 & 72.1 \\
                Anthropic Claude 2 (representative samples) & 1-shot & 76.8 & 75.1 \\
                
            \midrule

                MPNet-v2 (SetFit) & 3-shot & 76.7 & 75.9 \\
                GPT-3.5 (random samples) & 3-shot & 57.9 & 59.8 \\
                GPT-3.5 (representative samples) & 3-shot & 65.5 & 65.3 \\
                GPT-4 (representative samples) & 3-shot & \textbf{83.1} & \textbf{82.7} \\
                GPT-4 (random samples) & 3-shot & 74.2 & 73.7 \\
                \midrule
                MPNet-v2 (SetFit) & 5-shot & 83.5 & 83.3 \\
                MPNet-v2 (SetFit) & 10-shot & 88.1 & 88.1 \\
                MPNet-v2 (SetFit) & 15-shot & 90.6 & 90.5 \\
                MPNet-v2 (SetFit) & 20-shot & 91.2 & 91.3 \\
                
    \bottomrule
    \end{tabular}
\caption{Classification results for all models on the test data, with N-Shot indicating the number of samples used during training. The MPNet model is fine-tuned without the \texttt{SetFit} method on the Full-Data setting.}
\label{tab:classification_results}
\vspace{-1.1cm}
\end{table}

Despite being presented with only 1 sample per class (either random or representative), GPT-4 achieves competitive results (80.39 and 77.6 \microf). It outperforms MPNet-v2 by a large margin (over 20 pp) in the 1-shot setting, showing the potential for effective few-shot classification in domains where data is limited \cite{loukas-etal-2021-edgar}. 
Using our human-curated representative samples leads to better in-context learning results (both in GPT-3.5 and GPT-4). 
We also employed two alternative closed-source LLMs by Cohere and Anthropic on the representative samples. Cohere's Command-nightly  performs poorly with a low 58.4 \microf  while Anthropic's Claude 1 yields a 73.8 \microf, comparable to GPT-3.5's 75.2. At the same time, Anthropic's Claude 2 performs slightly better with a 76.8 \microf.

We proceed with 3-shot classification with the top-performing model, GPT-4, and the less powerful GPT-3.5 model of the OpenAI models family.
GPT-4 outperforms both previous models on the 3-shot setting by more than 6 pp (MPNet-v2) and 17 pp (GPT-3.5). More notably,
GPT-3.5, performs poorly on its 3-shot variant (65.5 \microf) compared to its promising 1-shot variant (75.2 \microf). This probably verifies recent reports on GPT-3.5 getting lost in the middle of long contexts \cite{liu2023lost}. Similarly to the 1-shot experiments, GPT-4's performance drops substantially (approximately 9 pp) when shown random samples instead of representative ones.
Thus, 
it is better to present more examples to a more powerful engine like GPT-4 instead of GPT-3.5.

\section{Cost-Effective LLM Inference using Retrieval-Augmented Generation}\label{cost-analysis}
\subsection{Cost Analysis}
Apart from studying the models' performance, 
we investigate the significant Operating Costs (OpEx) associated with popular LLMs like GPT-4. 
We provide a budget analysis for our experiments in Table ~\ref{tab:model-cost-analysis} (cost refers to $\sim$3k instances). This can be seen as a guideline for researchers and practitioners to evaluate the trade-offs between performance and budget when selecting a closed-source LLM offered via a managed API, as well as a data point when deciding between a ``build vs. buy'' approach, which requires developing datasets and setting up and hosting their own custom model.

\begin{table}[h]
    \centering
    \begin{tabular}{lccr}
        \toprule
        Model & Setting & Micro-F1$\uparrow$ & Cost$\downarrow$ \\
        \midrule
        GPT-4 & 1-shot & \textbf{80.4} & 620\$ \\
        GPT-3.5 & 1-shot & 75.2 & 31\$ \\
        Anthropic Claude 2 & 1-shot & 76.8 & \textbf{15\$} \\ 
        Command-nightly  & 1-shot & 58.4 & 22\$ \\
        \midrule 
        GPT-3.5 & 3-shot & 65.5 & \textbf{62\$} \\
        GPT-4 & 3-shot & \textbf{83.1} & 740\$ \\
        \midrule 
        \midrule 
        GPT-4 & 5 similar (RAG)  & 84.5 & 205\$ \\
        Anthropic Claude 2 & 5 similar (RAG) & \textbf{84.8} & \textbf{33\$} \\
        \midrule 
        GPT-4 & 10 similar (RAG) & 81.2 & 230\$ \\
        Anthropic Claude 2 & 10 similar (RAG) & \textbf{85.2} & \textbf{37\$} \\
        \midrule 
        GPT-4 & 20 similar (RAG) & \textbf{87.7} & 270\$ \\
        Anthropic Claude 2 & 20 similar (RAG) & 85.5 & \textbf{42\$} \\
        \bottomrule
    \end{tabular}
    \caption{Cost analysis of various closed-source LLMs. The first two groups represent the 1- and 3-shot results, as shown in Section \ref{sec:results} (Results). The rest groups come from Section ~\ref{sec:cost-effective-llm-inference}, where we utilize Retrieval-Augmented Generation (RAG) for cost-effective LLM inference. We perform 3,080 queries to each LLM, i.e., 1 query per sample in the test set.}
    \label{tab:model-cost-analysis}
  \vspace{-2mm}
    
\end{table}

Focusing on the two upper groups of the table, we see how popular LLMs perform in the financial intent detection task of Banking77 in a standard few-shot approach. We observe that in the 1-shot setting, Anthropic Claude 2 has a performance that comes on par with the GPT-3.5 model and comes at half of GPT-3.5's cost. In the 3-shot setting, GPT-4 scores nearly 20 points more than GPT-3.5, but comes with a 10x cost, which amounts to 740\$.

\subsection{Retrieval-Augmented Generation (RAG)}\label{sec:cost-effective-llm-inference}
Motivated by the high costs associated with closed-source LLMs \cite{2023-ocats}, 
we present a methodology to query LLMs efficiently by creating a Retriever component before plugging it into our Generative Text Classifier (Reader) pipeline. 

In the standard few-shot setting, we provide examples for all intents; e.g., in the 3-shot setting, we use a prompt with 231 samples (3 samples x 77 classes) as context in each inference call to the LLMs. However, our intuition is that a subset of them can be sufficient, allowing room for a cost-effective approach by narrowing down the number of representative examples used in each inference call.

\begin{figure}[h]
    \centering
    \includegraphics[width=0.85\columnwidth]{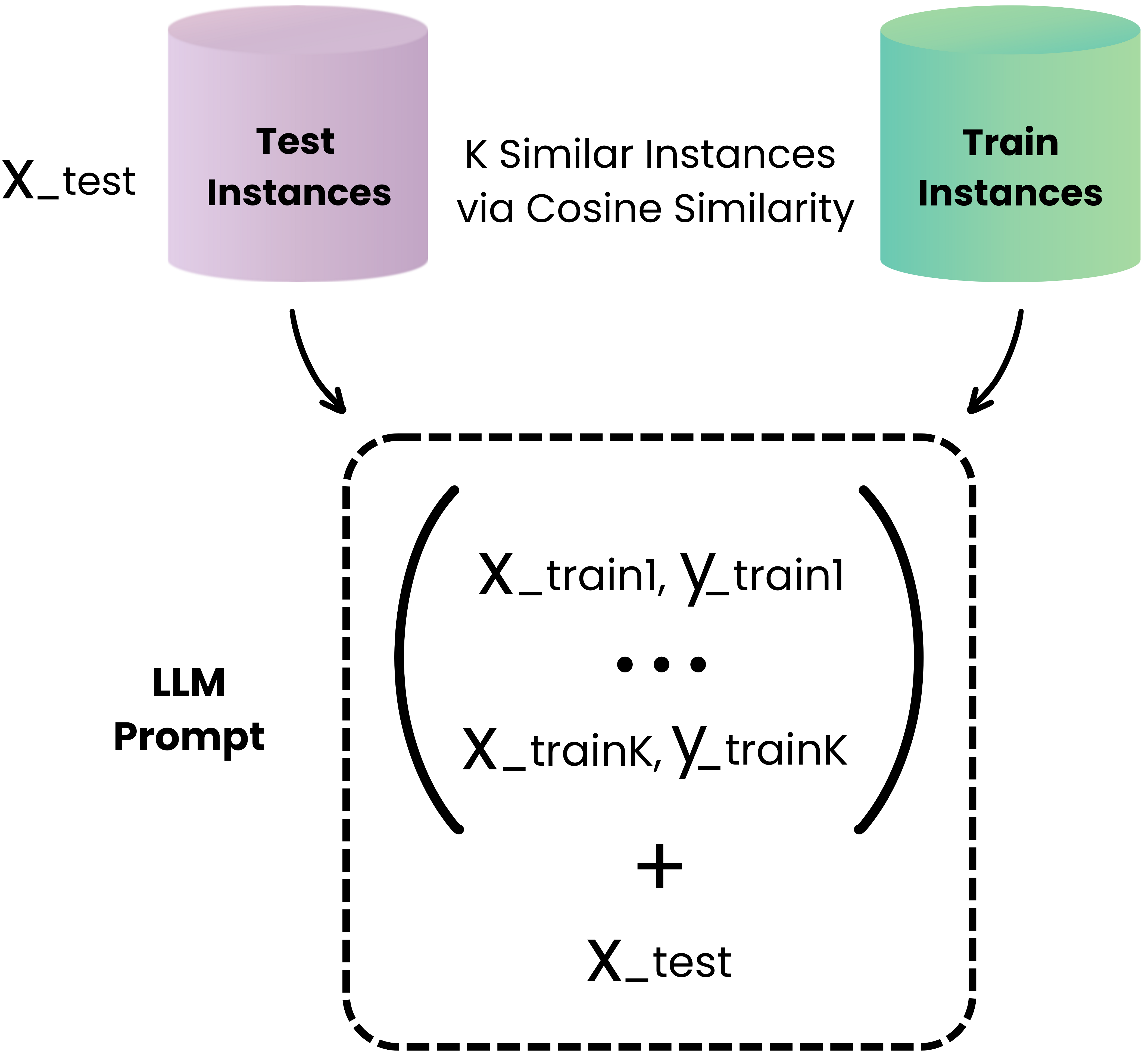}  
    \caption{Dynamic LLM prompt construction through Retrieval-Augmented Generation (RAG), using cosine similarity for in-context data selection. We use K=5, 10, 20.}  
    \label{fig:cost_effective_llm_inference_method}  
\label{fig:cosine-similarity}
\end{figure}

Thus, we consider including only the most similar examples to be utilized during each inference call. This is based on the active learning algorithms proposed by \citeauthor{rag2020} and \citeauthor{liu-etal-2022-makes}, who augment in-context examples for text generation language models using kNN. The rationale is: for each inference sentence to be classified, instead of providing all the 231 examples for the classes inside the prompt, we provide only the 5 (2.2\%), 10 (4.3\%), and 20 (8.7\%) most similar examples, as shown in Figure ~\ref{fig:cosine-similarity}. We use the \texttt{all-mpnet-base-v2} vector embeddings 
and the cosine similarity metric for the distance calculation.

The results of this approach, as seen in the three lower groups of Table \ref{tab:model-cost-analysis}, show improved results both in performance and budget.
For instance, when we retrieve only five similar instances (2.2\% of the 231 total examples) for each inference call, Anthropic Claude 2 has an 84.8 \microf vs. GPT-4's 84.5. At the same time, it costs 172\$ less (1/6 of GPT-4's cost). Most importantly, this score is also higher than 1.5 percentage points than GPT-4 in the classic 3-shot setting, which costs 740\$ and is 22 times multiple of its cost.

By increasing the retrieved similar samples to 10 and 20, we see a slight increase to a maximum of 87.7\% \microf for GPT-4 (270\$) and 85.5\% \microf of Claude 2 (42\$). Anthropic Claude 2 is consistent in improving over the number of retrieved examples, while GPT-4 shows a slight drop in performance when presented with 10 similar examples vs. 5 similar examples. 
One possible explanation for this is the non-deterministic behavior of LLMs.

\section{LLMs for Data Generation in Low-Resource Settings}
Data-centric AI is increasingly gaining attention, as evidenced by relevant scientific venues.\footnote{\url{https://datacentricai.org/neurips21/}} 
Such approaches become invaluable in areas where collecting vast quantities of data is either infeasible, prohibited by privacy constraints, or challenging such as in finance.\footnote{\url{https://sites.google.com/view/icaif-synthetic/home}}
They focus on maximizing the utility of limited datasets rather than only focusing on scaling up model complexity (model-centric AI), which is often impractical to maintain or deploy \cite{emc2-2019-energy-efficiency, knowledge-distillation}.

In line with this low-resource paradigm, we simulate a scenario where our dataset is limited. We assume that providing 3 examples per class is feasible in practice, we start with the 231 original handpicked examples from the expert (3 per each of the 77 classes), and we leverage GPT-4 to generate additional examples per class. The intention here is to explore how much we can rely to data augmentation to increase the performance, and identify the threshold beyond which the quality of generated examples starts to degrade, and thus, we cannot continue with adding more (artificial) examples.

While past existing studies show that just writing simple prompts boost performance on various classifications tasks, they come short due to two reasons: (a) they consider vanilla tasks with up to 6 classes (where in our scenario, we consider 77) \cite{yoo-etal-2021-gpt3mix-leveraging} and (b) they use older text generation models rather than instruct-based LLMs.

Since \citeauthor{sahu-etal-2022-data-servicenow} found that data augmentation for tasks with large label volume is likely to not benefit at all (at least when using older text generation models), we adopt a particular prompt template for our instruct-based GPT-4 model. First, we analyze the data using Azimuth \cite{gauthier-melancon-etal-2022-azimuth}, an open-source toolkit. We split the 77 labels into ten groups, where each group consists of labels that contain intents with highly semantic overlap, as shown in Azimuth (e.g., Top Up Reverted and Top Up Failed belong to the same group). Then, we prompted the model with the 3 examples of each class inside that group, and we specifically requested it to generate 20 artificial examples by paying attention to these confounding classes.\footnote{Our prompt template for data generation can be found in \url{https://anonymous.4open.science/r/data-prompt-4805}.}

Subsequently, we employed the SetFit approach across four different settings: 5-shot, 10-shot, 15-shot, and 20-shot. However, to incorporate the generated data, we amended the traditional methodology of presenting the N original examples per class. 
Instead of the 5-shot experiment, we used 3 examples from the original data and 2 from the augmented. Similarly, the original-to-augmented ratios for 10-shot, 15-shot, and 20-shot tasks were 3:7, 3:12, and 3:17, respectively. This allowed us to assess how well our model performs with limited data and how effectively it can integrate and learn from artificially generated examples.

\begin{figure}[t]
    \centering
    \includegraphics[width=1.0\columnwidth]{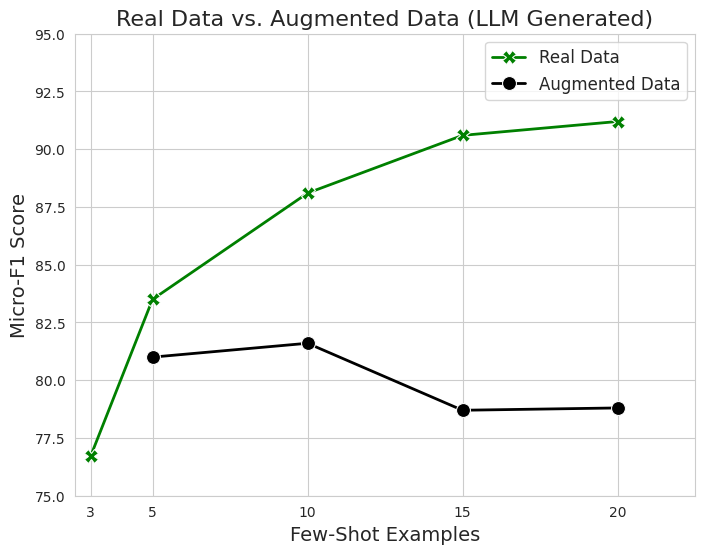}  
    \caption{The Micro-F1 Score for various few-shot settings. In the Augmented Data (black) line, 3 of the examples each time (out of the 5, 10, 15, 20) belong to the representative samples that the human expert picked from the real data.} 
    \label{fig:real_data_vs_generated_data}  
  \vspace{-2mm}
\end{figure}

Figure ~\ref{fig:real_data_vs_generated_data} shows the \microf score for different few-shot settings. Starting from the 3-shot setting with 76.7\% \microf, one can generate artificial data to get an essential boost to 81\% \microf as seen in the augmented 5-shot setting (3 original examples + 2 generated). The maximum performance increase is observed at the 10-shot augmentation scenario with 81.6\%. After this threshold, the quality of the data generated decreases (see 15- and 20-shot settings), injecting more noise into the model than the actual value. As expected, the real data (even if they are challenging to obtain in such domains) are far superior and more meaningful than the generated data (91.2 \microf vs. 78.8 \microf in the 20-shot setting).

\section{Error Analysis}\label{sec:error-analysis}
Lastly, to understand our models' limitations, we also performed an exploratory error analysis in some of our top-performing models.
Specifically, we manually inspected the errors of GPT-3.5 (1-shot), GPT-4 (3-shot), MPNet-v2 (10-shot), and MPNet-v2 (Full-Data).

\subsection{Errors in LLMs}

The most misclassified labels by GPT-4 (3-shot) and GPT-3.5 (1-shot) are shown in Table ~\ref{tab:misclassifications_in_gpt_35_gpt_4} along with their percentages.
After inspecting the samples and their predicted labels, we observe that the label \texttt{Get Physical Card} was frequently misclassified as \texttt{Change Pin}.
This is potentially due to the presence of the word \textit{``PIN''} in all test instances, causing the vector embeddings to locate them close to the decision boundary of the \texttt{Change Pin} class. As for the \texttt{Transfer Not Received By Recipient} class, phrases indicative of non-receipt were observed in only 20\% of the test samples, which could have complicated the classification task. The rest of the samples exhibited concerns related to transaction timing and recipient visibility, pushing them towards the \texttt{Transfer Timing} class.

\begin{table}[h]
  \centering
  
  \begin{tabular}{lcc}
        \toprule
        \multirow{2}{*}{Gold Label} & \multicolumn{2}{c}{Misclassifications} \\
        \cmidrule{2-3}
         & GPT-4 & GPT-3.5 \\
        \midrule
    Get Physical Card & 87.5\% & 87.5\% \\
    Transfer Not Received & 62.5\% & 50.0\% \\
    Beneficiary Not Allowed& 42.5\% & 60.0\% \\
    \bottomrule
  \end{tabular}
  \caption{Top misclassified labels, along with their misclassification percentages (out of 40 test instances), for the GPT-4 (3-shot) and GPT-3.5 models (1-shot).}
  \label{tab:misclassifications_in_gpt_35_gpt_4}
  \vspace{-2mm}
\end{table}
The third label presented in the table, \texttt{Beneficiary Not Allowed}, was incorrectly classified mostly as \texttt{Declined Transfer} by GPT-4 and GPT-3.5, respectively. The incorrect labeling could be attributed to sentences that might seem ambiguous even for a human classifier. For instance, sentences such as \textit{``I tried to make a transfer, but it was declined.''} and \textit{``The system does not allow me to transfer funds to this beneficiary.''} were misinterpreted as pertaining to \texttt{Declined Transfer} class, due to the mention of terms like \textit{``declined''} and \textit{``does not allow''}. The models, thus, failed to discern the actual issue of beneficiary disallowance.

\subsection{Errors in MLMs}



In Table ~\ref{tab:full_data_vs_10shot}, the MPNet-v2 model displayed significant confusion between in two labels when trained on Full-Data and 10 samples per class. Specifically, after inspection, we see that for the gold label \texttt{Top Up Failed}, the sentence \textit{``please tell me why my top-up failed''} was misclassified as \texttt{Top Up Reverted}, likely due to contextual overlap between failed and reverted top-ups. Another source of confusion involved the \texttt{Transfer Timing} gold label. Sentences like \textit{``How many days does it take until funds are in my account?''} were misclassified as \texttt{Pending Transfer} or \texttt{Balance Not Updated After Cheque or Cash Deposit}. The temporal aspect of the query seems to direct the models towards other classes that also involve time, thus highlighting challenges in differentiating specific concerns related to money transfers.

\begin{table}[h]
    \centering
    \begin{tabular}{lcc}
        \toprule
        \multirow{2}{*}{Gold Label} & \multicolumn{2}{c}{Misclassifications} \\
        \cmidrule{2-3}
         & Full-Data & 10-shot \\
        \midrule
        Top Up Failed & 22.5\% & 17.5\% \\
        Transfer Timing & 20.0\% & 45.0\% \\
        \bottomrule
    \end{tabular}
    \caption{Misclassification percentages (out of 40 test instances) for the MPNet-v2 model when trained on Full-Data vs. when trained on 10 samples per class.}
    \label{tab:full_data_vs_10shot}
  \vspace{-6mm}
\end{table}




\subsection{About Overlapping Labels}
We anticipate that this analysis will serve as a source of inspiration for future work in financial intent detection. One possible approach to mitigate these errors involves the adoption of hierarchical classifiers \cite{chalkidis-etal-2020-empirical}. These classifiers could initially identify classes within broad categories, such as transfers or top-ups, and subsequently classify them into more specific classes, such as pending transfer vs. transfer timing and top-up failed vs. top-up reverted. We believe that these methodologies could either be integrated with ``native'' TensorFlow/PyTorch pipelines or simulated using  Chain-of-Thought prompting techniques \cite{wei2022chain} so the LLMs can distinguish fine-grained financial labels by thinking step-by-step or class-by-class.

\section{Conclusion}

We conducted a comprehensive few-shot text classification study using LLMs, MLMs and discussed their trade-offs between cost and performance. We focused on Banking77, a financial intent detection dataset with real-life challenges, such as its large number of intents and overlapping labels. 

Our results clearly demonstrate the effectiveness and efficiency of in-context learning using conversational LLMs. This approach serves as a practical and rapid solution for achieving accurate results in few-shot scenarios and domains with restricted resources, like finance.
In detail, we demonstrated that LLMs, like GPT-3.5, GPT-4 and Anthropic Claude 2, can perform better than MLMs in some limited-data scenarios (1- and 3-shot). 
On the other side, by fine-tuning custom MLMs like MPNet-v2 with SetFit, we surpassed the previous work of \cite{mehri-eric-2021-example} in the 10-shot setting by 2.2 pp. 

LLM services reduce the need for technical skills and skip GPU training times. However, they can be considered costly for small organizations, given that they are accessed behind paywalls (740\$ for 3-shot experiments of $\sim$3k samples with GPT-4).
After presenting detailed pricing costs to help the community make better decisions, we also demonstrated a cost-effective inference method for LLMs. We utilized a semantic similarity retriever to return only a small but substantial fraction of training examples to our prompt, showing that this performs better than classic few-shot in-context learning. Most importantly, using Anthropic Claude 2 with this RAG approach, we can save multiple (22) times the cost associated with LLM services like GPT-4 (700\$ less) while getting a higher classification score.

To showcase how financial companies can utilize LLMs in scenarios with limited data, we also used GPT-4 to generate artificial data for data augmentation. We conclude that these generated data points are helpful up to a specific threshold (7 generated examples), after which our performance starts to drop, possibly due to the LLM starting to generate noise. 

Lastly, all of our top-performing Few-Shot experiments using LLMs and MLMs were trained on representative data samples out of a human expert-curated Banking77 subset. We provide this curated dataset freely available in order to promote the development of robust financial AI systems.

\section{FUTURE WORK}
In future work, we plan to experiment with open-sourced LLM alternatives, which may be suitable substitutes for closed-source models, like LLaMA2 \cite{touvron2023llama2} and incorporating Chain-of-Thought \cite{wei2022chain} techniques to mitigate the errors about the overlapping class labels, as presented extensively in the error analysis section.

\begin{acks}
This work has received funding from European Union's Horizon 2020 research and innovation programme under grant agreement No 101021714 ("LAW GAME"). Also, we would like to sincerely thank the Hellenic Artificial Intelligence Society (EETN) for their sponsorship.
\end{acks}

\bibliographystyle{ACM-Reference-Format}
\bibliography{sample-base}

\clearpage

\end{document}